\def\year{2022}\relax
\documentclass[letterpaper]{article} % DO NOT CHANGE THIS
\usepackage{aaai22}  % DO NOT CHANGE THIS
\usepackage{times}  % DO NOT CHANGE THIS
\usepackage{helvet}  % DO NOT CHANGE THIS
\usepackage{courier}  % DO NOT CHANGE THIS
\usepackage[hyphens]{url}  % DO NOT CHANGE THIS
\usepackage{graphicx} % DO NOT CHANGE THIS
\usepackage{natbib}  % DO NOT CHANGE THIS AND DO NOT ADD ANY OPTIONS TO IT
\usepackage{caption} % DO NOT CHANGE THIS AND DO NOT ADD ANY OPTIONS TO IT
\DeclareCaptionStyle{ruled}{labelfont=normalfont,labelsep=colon,strut=off} % DO NOT CHANGE THIS
\usepackage{amsmath}
\usepackage{booktabs}
\usepackage{xcolor}
\usepackage{subfigure} 
\usepackage{algorithm}
\usepackage{algorithmic}

\newcommand{\figref}[1]{Fig. \ref{#1}}
\newcommand{\tabref}[1]{Tab. \ref{#1}}

\usepackage{newfloat}
\usepackage{listings}
\floatstyle{ruled}
\newfloat{listing}{tb}{lst}{}
\floatname{listing}{Listing}
\title{Receptive Field Broadening and Boosting for Salient Object Detection}
\author{
    Mingcan Ma\textsuperscript{\rm 12},
    Changqun Xia\textsuperscript{\rm 2$*$},
    Chenxi Xie\textsuperscript{\rm 12},
    Xiaowu Chen\textsuperscript{\rm 1},
    Jia Li\textsuperscript{\rm 12}\footnote{Correspondence should be addressed to Changqun Xia and Jia Li. URL: http://cvteam.net.}
}
\affiliations{
    \textsuperscript{\rm 1}State Key Laboratory of Virtual Reality Technology and Systems, SCSE, Beihang University, Beijing, China\\
    \textsuperscript{\rm 2}Pengcheng Laboratory, Shenzhen, China
}

\usepackage{bibentry}
\begin{document}

\maketitle

\begin{abstract}
Salient object detection requires a comprehensive and scalable receptive field to locate the visually significant objects in the image. Recently, the emergence of visual transformers and multi-branch modules has significantly enhanced the ability of neural networks to perceive objects at different scales. However, compared to the traditional backbone, the calculation process of transformers is time-consuming. Moreover, different branches of the multi-branch modules could cause the same error back propagation in each training iteration, which is not conducive to extracting discriminative features. To solve these problems, we propose a bilateral network based on transformer and 
CNN to efficiently broaden local details and global semantic information simultaneously. Besides, a Multi-Head Boosting (MHB) strategy is proposed to enhance the specificity of different network branches. By calculating the errors of different prediction heads, each branch can separately pay more attention to the pixels that other branches predict incorrectly. Moreover, Unlike multi-path parallel training, MHB randomly selects one branch each time for gradient back propagation in a boosting way. Additionally, an Attention Feature Fusion Module (AF) is proposed to fuse two types of features according to respective characteristics. Comprehensive experiments on five benchmark datasets demonstrate that the proposed method can achieve a significant performance improvement compared with the state-of-the-art methods.
\end{abstract}
\begin{figure}[t]
\centering
\includegraphics[width=0.99\columnwidth]{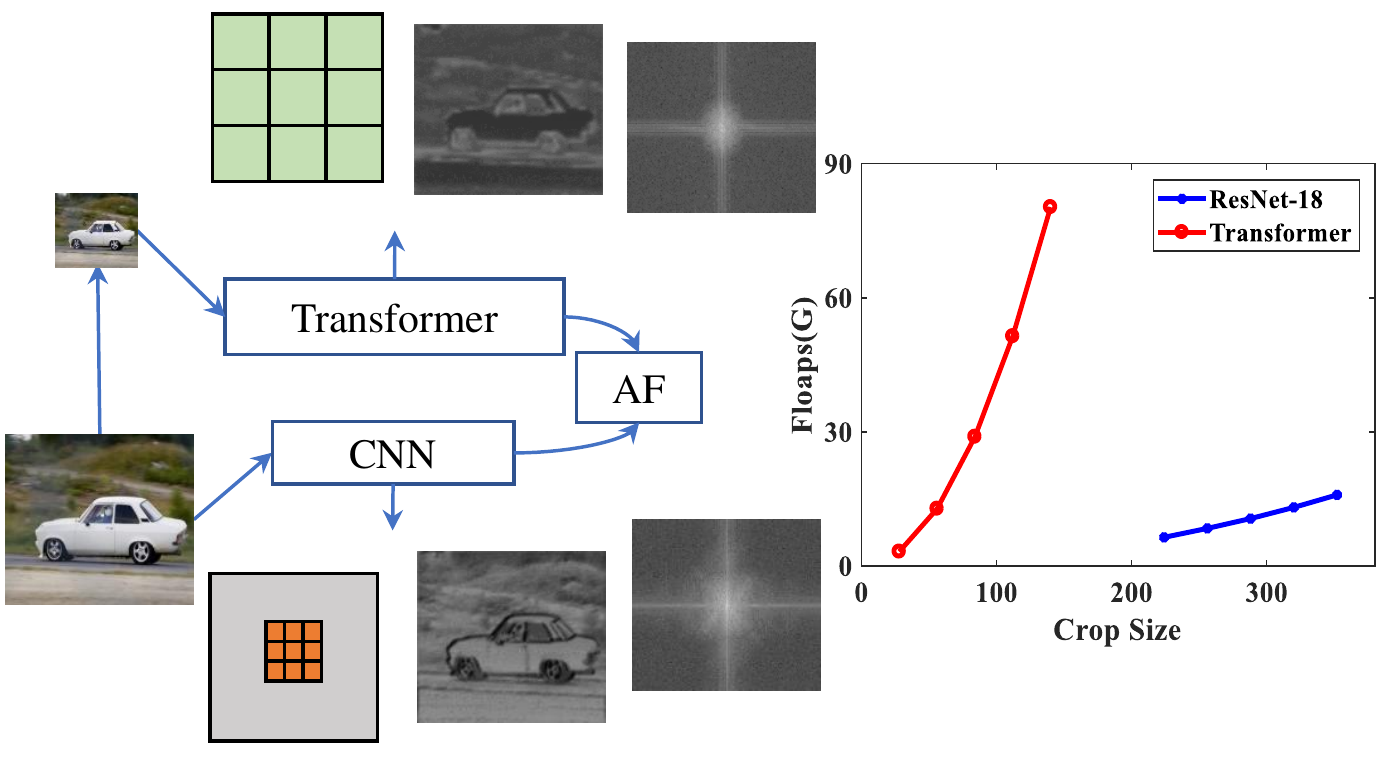} % Reduce the figure size so that it is slightly narrower than the column. Don't use precise values for figure width.This setup will avoid overfull boxes.
\caption{Comparison of the two types of models. From left to right, it shows the receptive field, features, frequency domain signals and efficiency. ResNet-18 \cite{ResNet} and Swin-Transformer \cite{liu2021swin} are selected as the basic network to extract details and deep semantics respectively.}
\label{fig1}
\end{figure}

\section{Introduction}
Recently, CNN-based methods have made great progress in the SOD field due to the powerful feature representation. However, due to the limitations of the convolutional receptive field, it is difficult for the CNN networks to break through the existing performance. The emergence of visual transformers has broken the current visual attention mechanism and can understand images from an overall perspective. However, the price of a larger receptive field is more computational cost, especially when the input image resolution is larger.

Actually, CNN and transformers have their own distinctiveness. As shown in Fig. 1, we observe that: 1. Features are different in the frequency domain. CNN features reveal mostly high frequency signals, while transformer features contain mostly low-frequency signals. 2. The computational cost increases at different rates as the input resolution increases. Two types of models can both run fast on low-resolution images, but when the resolution is large, the calculation cost of Transformer is much higher than that of CNN. 3. There are differences in regard to the receptive field mechanism. CNN expands the receptive field from the local to the whole image through a bottom-up calculation method, while Transformer obtains global information directly from the overall perspective.

Based on the observation of these distinctiveness, we further analyzed the feasibility of complementarity between CNN and transformers from three aspects: 1. The signal difference in frequency domain indicates that the feature characteristics of the two models can be complementary. 2. The processing difference in attention mechanism reflects that their ways of understanding images can be complementary. 3. The difference in processing speed reveals that the input sizes can be complementary.
Therefore, we consider using a transformer to extract global semantic information and CNN to extract detailed information together. In this way, transformer does not require a large input size and CNN does not require a deeper convolutional network, which can achieve a better balance of computational costs and performance. To this end, we propose to build an effective bilateral network to balance two kinds of requirements. The transformer model with low-resolution input is used to obtain low-frequency semantic information, and the lightweight CNN with high-resolution input is used to analyze high-frequency details. It is worth noting that the characteristic of the CNN models is that they can establish a perfect channel connection but the spatial receptive field is insufficient, while the transformer models can establish a full-space connection but the channel connection is relatively weak. So, Attention Fusion Module (AF) is proposed to explore a better integration strategy of two types of features.

 \begin{figure}[t]
\centering
\includegraphics[width=0.99\columnwidth]{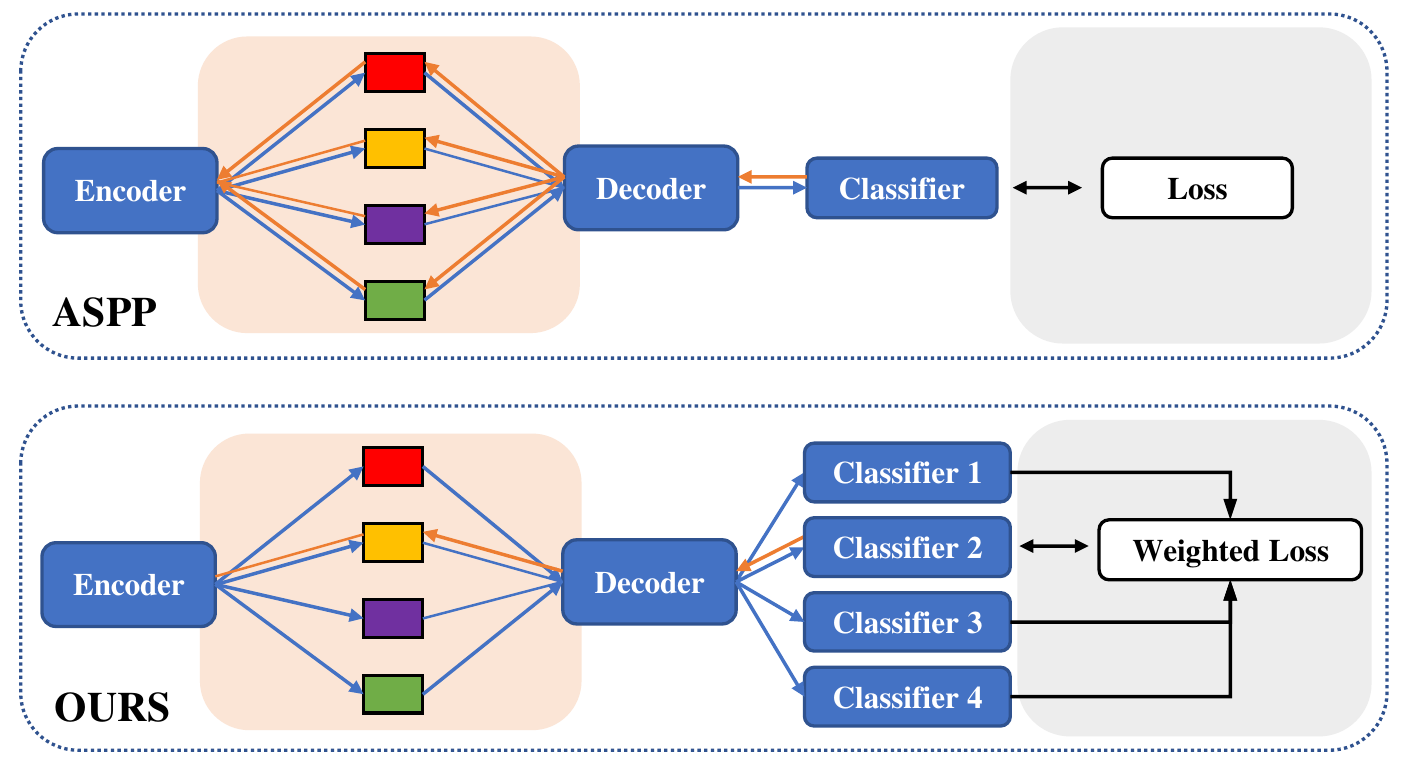} % Reduce the figure size so that it is slightly narrower than the column. Don't use precise values for figure width.This setup will avoid overfull boxes.
\caption{A logical comparison diagram between the proposed Multi-Head Boosting and ASPP \cite{ASPP}. The colored squares represent dilated convolutions with different atrous rates. The blue arrow represents forward propagation, and the orange arrow represents the only branch randomly selected during back propagation.}
\label{fig2}
\end{figure}
However, when the input resolution becomes smaller, the transformer’s global attention mechanism degenerates into an association between regions, which cannot capture enough signal range. Therefore, the goal of increasing the receptive field cannot be totally dependent on the transformer branch. To enhance the receptive field of CNN across regions, multiscale modules such as ASPP (Chen et al. 2017) and PPM (Zhao et al. 2017) are good choices due to the strong ability of extracting features of different scales in different branches. But remarkably, the current multi-branch module adopts the strategy of parallel training and synchronous error back propagation. This method is not conducive to extracting discriminative features separately for different branches. 

Therefore, we put forward a novel strategy named Multi-Head Boosting (MHB) as shown in Figure 2. MHB mainly has the following improvements: 1. Forward propagation activates all paths, and the error back propagation process only activates a random branch. 2. A separate prediction head is set for each branch, and each branch is back-propagated by pixel weighting to focus on the error parts of other branches. With MHB, we can effectively construct the association between different regions. The combination of bilateral network and MHB can effectively enhance the receptive field of the network and thus achieve higher SOD performance. 

Finally, we design ablation experiments to verify the effectiveness of the proposed bilateral framework and multi-head boosting strategy. The results of comprehensive comparison experiments show that our method has greatly exceeded the current state-of-the-art methods. Our main contributions can be summarized as follows:

\begin{itemize}
\item We propose to construct an effective bilateral network based on CNN and transformer. The lightweight CNN is responsible for fast detail extraction at high resolution, and the transformer model can efficiently generate global correlation features at low resolution input. 
\item We propose an attention feature fusion module (AF) to fuse the features of the two types of models. AF considers the advantages of the two features in space and channel dimensions to strengthen each other.
\item
We propose a novel multi-head boosting strategy, which makes each branch pay more attention to the pixel position of each other's prediction error and breaks the synchronization relationship of the error back-propagation process.
\item The proposed method achieves a great performance improvement over the state-of-the-art methods. In particular, The MAE metric of our method reach $\textbf{0.021}$ on ECSSD and $\textbf{0.040}$ on DUT-OMRON, which is $\textbf{29\%}$ and $\textbf{20\%}$ lower than the previous methods. 
\end{itemize}
\begin{figure*}[t]
\centering
\includegraphics[width=0.98\textwidth]{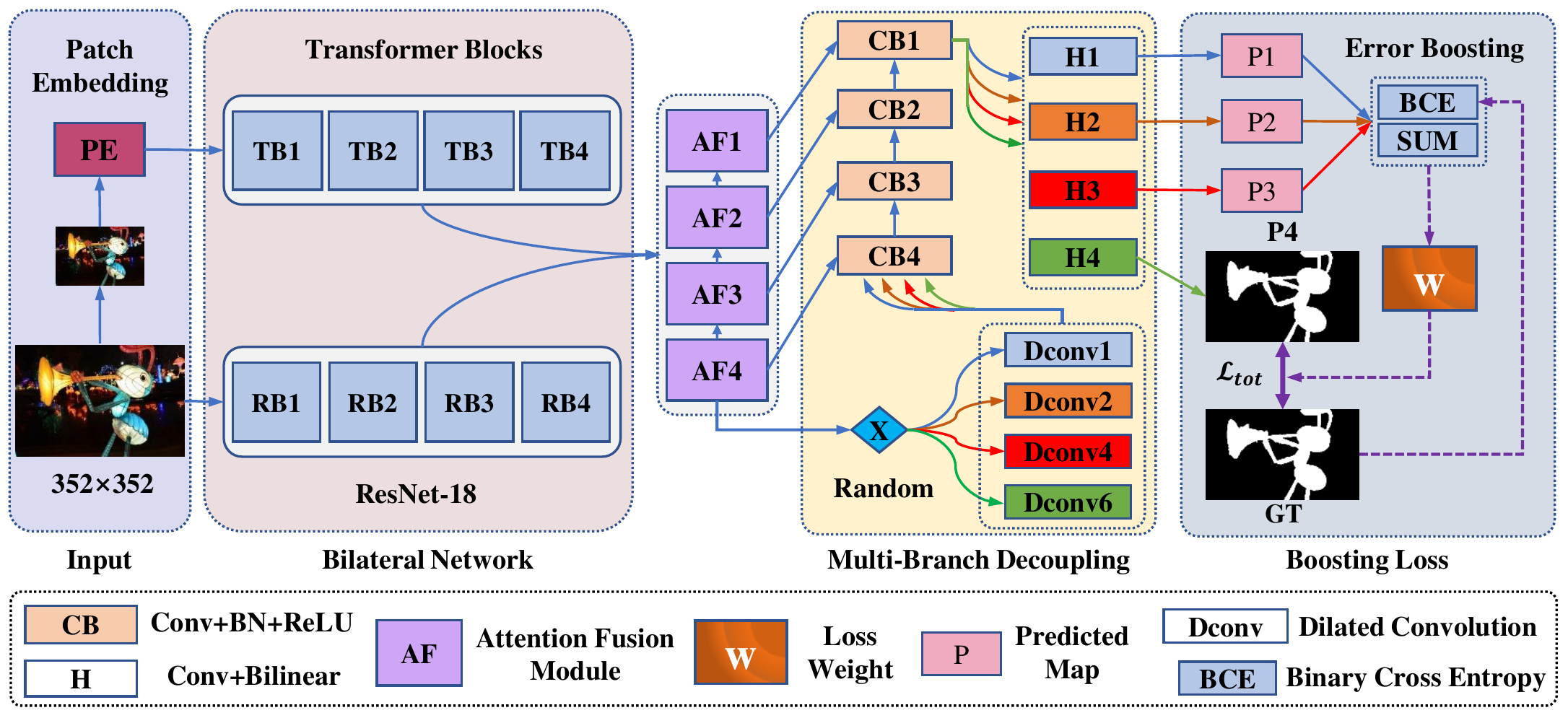} % Reduce the figure size so that it is slightly narrower than the column.
\caption{Framework diagram of the proposed method. This figure shows the training process where the random number $X$ takes a value of $4$. The feature fusion module in the figure accepts the input of the bilateral module with the corresponding number. The prediction result of $H_{4}$ is used for gradient back propagation, other branches do not calculate the gradient, and their results are used to calculate the error weight.}
\label{fig3}
\end{figure*}

\section{Related Work}
\subsection{Efficient Two-Stream Network Designs}
Computer vision tasks often need to extract detailed information and semantic information of an image. Detailed information does not depend on the depth of the network but requires larger input resolution, while semantic information is just the opposite. In order to balance these two processes, the dual-branch network is proposed to obtain these two features separately, such as BiSeNet \cite{BiSeNet}. This method can better balance efficiency and performance. However, the backbones used by the two branches are often similar convolutional neural networks, which often brings a lot of redundancy. For example, StdcNet \cite{Rethink-BiSe} propose that better results can be obtained without a separate detail branch. In fact, this two-stream fusion idea is not only suitable for the fusion between convolutional neural networks. Recently, more and more visual backbones have been proposed, such as ResNet \cite{ResNet}, MobileNet \cite{mobilenets}, StdcNet \cite{Rethink-BiSe}, ShuffleNet \cite{zhang2018shufflenet}, Swin Transformer \cite{liu2021swin}, T2T-VIT \cite{t2tvit}, etc. These models contain efficient lightweight CNN and high computational cost networks based on visual self-attention. Compared with fusing different convolutional neural networks, we propose to fuse CNN and transformer frameworks with different but complementary mechanisms.
\subsection{Multi-scale perception}The multi-scale perception ability of neural networks often depends on the range of the receptive field. The multi-branch module is an effective method to enhance the receptive field of convolutional neural networks. For example, Pyramid Pooling Moudle \cite{PPM} and Atrous Spatial Pyramid Pooling \cite{ASPP} enhance the local scale perception ability of the network by designing multi-branch modules with different receptive fields. PoolNet \cite{Liu2019PoolSal} has used PPM for SOD for the first time and achieved good results. BaNet \cite{su2019banet} proposes to improve ASPP and achieve the latest new capabilities at the time. The above methods can effectively improve the expressive ability of the network, but there are still shortcomings.
The original intention of the multi-branch model is to allow different branches to capture specific features. However, most of the previous methods use a synchronous error back propagation strategy and seldom consider the specificity of the branch separately. In contrast, We use a similar multi-model boosting idea to achieve mutual enhancement of multiple branches. Each branch calculates the error separately and weights the error areas of the other branches.
\subsection{Accurate Detection of Salient Objects}In the past three years, the upper limit of SOD performance has been continuously broken by the latest methods. For example, F3Net \cite{wei2019f3net} proposes Fusion, Feedback and Focus to detect salient objects in 2019 and has achieved the best performance at the time. The MAE reach 0.053 on the DUT-OMRON dataset. In 2020, LDF \cite{LDF} decouples pixels based on the distance from the edge and iteratively optimizes the predicted maps, and its MAE on the same dataset reaches 0.051. In 2021, PA-KRN \cite{PAKRN} proposes a strategy of positioning first, then segmentation, and its MAE reaches 0.050. VST \cite{vst} uses the transformer framework for SOD for the first time, and its MAE reach 0.058. From the above results, it can be seen that the performance of SOD is getting better and better, but the improvement is very slow. We explore the differences between CNN and transformers in terms of features, computational complexity, and attention mechanisms. Based on this, we design a bilateral network and an Attention Feature Fusion Module to balance the two models and propose a Multi-Head Boosting method to compensate for the lack of association between regions. Due to the complementary combination, our method has achieved good results in both performance and efficiency. In particular, the MAE metric of our method on ECSSD has reached $\textbf{0.040}$.

\section{Proposed Method}In this section, we will introduce the details of the composition of the bilateral network, the Attention Feature Fusion Module, and the Multi-Head Boosting strategy. The training process of the model is shown in \figref{fig3}. The verification process only needs to add up the prediction results of each branch and then pass the activation function.
\subsection{Bilateral Network}As can be seen from \figref{fig1}, CNN and transformers have both  distinctiveness and complementarity. Therefore, we propose a bilateral network as shown in \figref{fig3} to fully utilize the complementary advantages. The bilateral network contains a semantic branch and a detail branch. The semantic branch makes full use of the global view of the transformer model to extract global semantic features and enhance the receptive field of the overall network. The detail branch is responsible for extracting high frequency details. The ingenuity of the bilateral network is that it can take advantage of the two types of models at the lowest computational cost. The complex semantic branch is mainly responsible for extracting global attention information so it can be calculated under low-resolution input. The detail branch only requires a lightweight CNN as basic network. Therefore, the combination of transformer and CNN can efficiently balance semantics, details, and their calculation cost.
\begin{figure}[t]
\centering
\includegraphics[width=0.97\columnwidth]{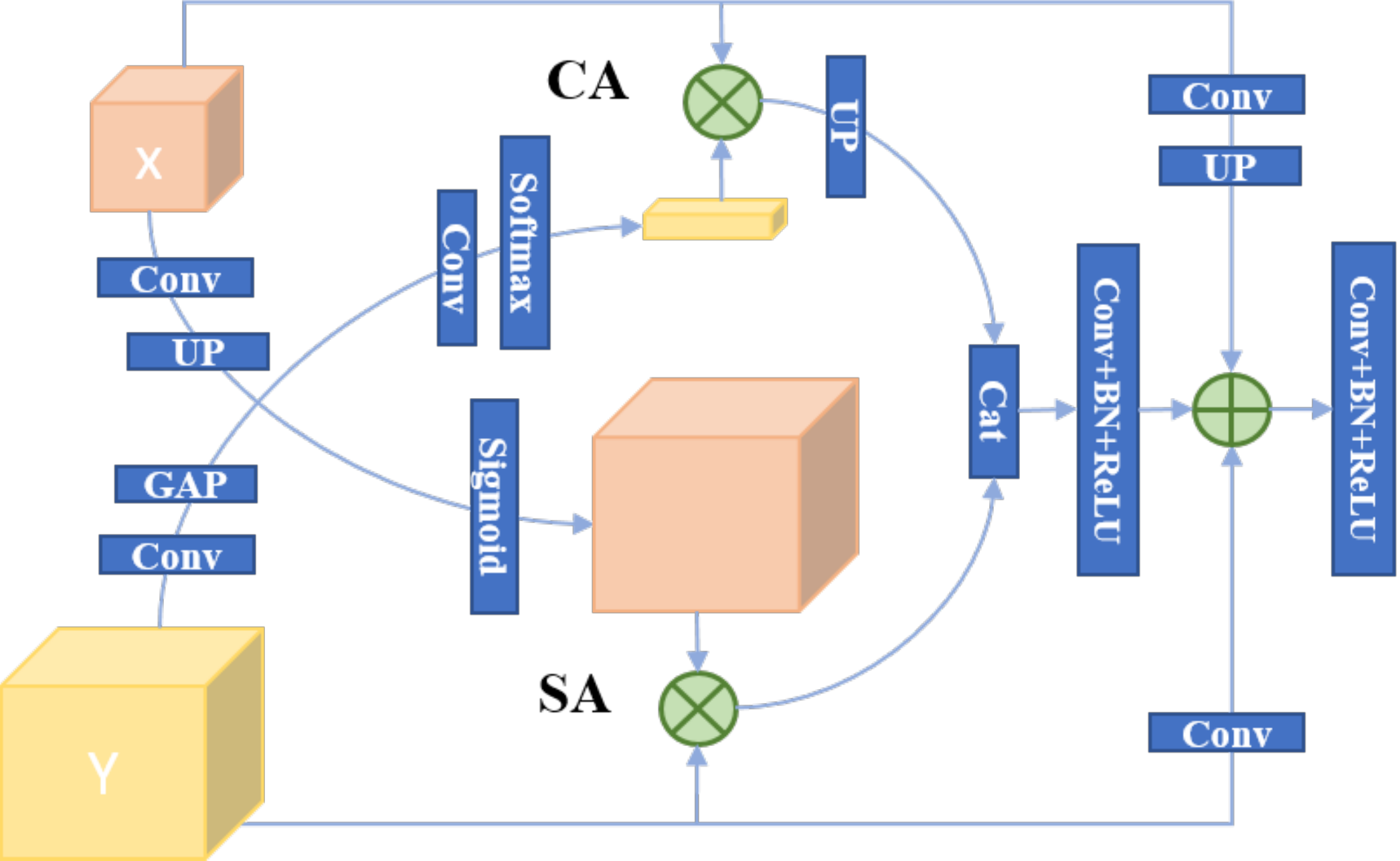} % Reduce the figure size so that it is slightly narrower than the column. Don't use precise values for figure width.This setup will avoid overfull boxes.
\caption{Illustration of the Attention Feature Fusion Module. X and Y represent the output feature of the transformer model and the output feature of CNN, respectively.}
\label{fig4}
\end{figure}
Specifically, we choose ResNet-18 \cite{ResNet} to build the detail branch. In order to enrich the detailed information extracted by this branch, we remove the first down-sampling operation to obtain more detailed features at high resolution. In this way, the resolution of the features output by each layer is $1/2$, $1/4$, $1/8$, and $1/16$ of the initial resolution. As for the semantic branch, we utilize Swin Transformer \cite{liu2021swin} to extract global features. This branch first samples the image to a resolution of $56\times 56$, and then passes through the self-attention modules of the transformer in turn. 

The working process of proposed bilateral network can be summarized as follows: Input the $352\times 352$ resolution image into the lightweight CNN to obtain multi-level detailed features $R_{1}$,$R_{2}$,$R_{3}$ and $R_{4}$. Besides, the input image is down-sampled to a resolution of $56\times 56$. Then we input it to the semantic branch to obtain features $T_{1}$, $T_{2}$, $T_{3}$, and $T_{4}$ and reshape the features. Finally, Use the attention feature fusion module $AF_{i}$ to fuse $T_{i}$ and $R_{i}$, where $i\in \{1, 2, 3, 4\}$.

\subsection{Attention Feature Fusion Module}
Unlike the features of CNN, the features of transformer model are composed of vectors stretched by pixel blocks. Given the transformer features, we first restore the previous positional relationship of the vector to get feature $X$ of \figref{fig4}. The transformer model calculates the correlation between all patch blocks in the space through the vector inner product, while CNN establishes the connection between all channels in the local space. The former constructs better spatial correlation, while the latter has stronger channel correlation. In order to better integrate the two types of features, we design an Attention Feature Fusion Module (AF). AF uses the respective characteristics of the two types of features to enhance the expression of each other. The specific process can be expressed as:
\begin{align}
F_{mid} &= \mathcal{C}_{br}(\mathcal{U}(\mathcal{R}_{ca}(Y)\otimes X)\textcircled{c}(\mathcal{R}_{sa}(X)\otimes Y)),\\
F_{out} &= \mathcal{C}_{br}(\mathcal{U}(\mathcal{C}(X))\oplus F_{mid} \oplus \mathcal{C}(Y)),
\end{align}
where $X$ and $Y$ correspond to the features in \figref{fig4}. $\mathcal{R}_{sa}$ and $\mathcal{R}_{ca}$ represent the operations corresponding to the crossed arrows in the figure. $\mathcal{U}$ represents the upsampling operation. $\mathcal{C}$ represents convolution, and the subscript $br$ represents a Batch Normalization and a ReLU activation. $\oplus$, $\otimes$ represent pixel-wise addition and multiplication operations, respectively. $\textcircled{c}$ denotes feature concatenation operation. $F_{mid}$ is the intermediate result, $F_{out}$ is the final output.

Specifically, the feature $X$ obtained by transformer contains rich semantic information, while the low level details are lost. CNN feature $Y$ is obtained by convolution, the spatial receptive field is smaller, but the correlation between channels is stronger. Therefore, we use the features of $X$ to select the spatial Features of $Y$, and use the channel information of $Y$ to enhance $X$. Finally, the final result is obtained by fusing all the information through the residual structure.

\subsection{Multi-Head Boosting}
The design idea of Multi-Head Boosting (MHB) is to improve the overall effect of the multi-branch module by decoupling the training process and adding branch complementary losses. It can be roughly divided into two parts: multi-path decoupling and boosting loss. Next, we will introduce these two processes in detail.

Multi-path decoupling aims to decouple multi-branch modules trained in parallel into multiple models. The significance is that the training process of different branches can be isolated, thereby reducing the gradient correlation between each branch and enhancing the scale specificity of different branches. ASPP integrates multi-channel features in the module, single-channel training cannot be performed directly, therefore we cannot use multi-channel features in the testing process. Our solution is to delay feature merging to the final result layer. In this way, the purpose of single-channel training and multi-channel testing can be realized.
\begin{figure*}[t]
\centering
\includegraphics[width=0.98\textwidth]{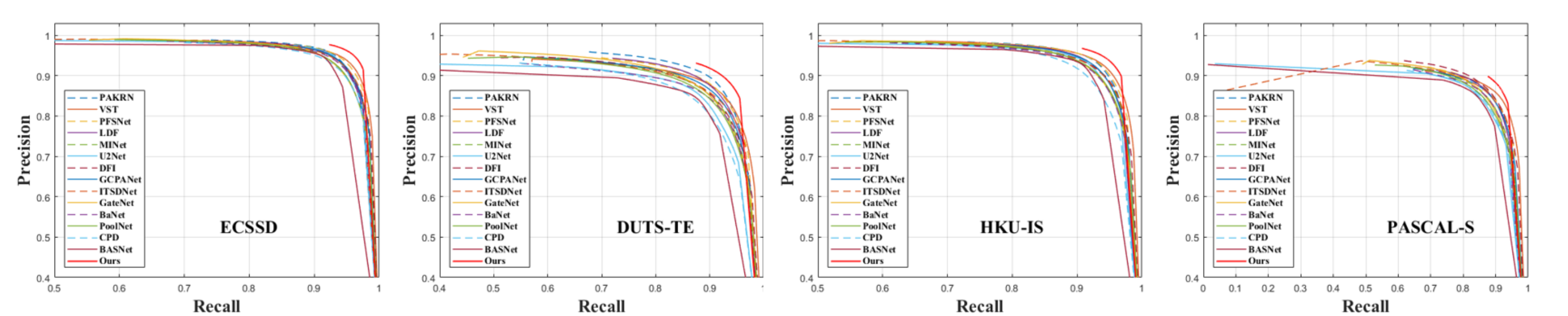} % Reduce the figure size so that it is slightly narrower than the column.
\caption{Comparison of PR curves between our method and the 14 latest methods.}
\label{fig5}
\end{figure*}
%\begin{figure*}[!ht]
%\centering
%\subfigure{
%\begin{minipage}[t]{0.235\linewidth}
%\centering
%\includegraphics[width=4.7cm]{111.eps}
%
%\end{minipage}
%}
%\subfigure{
%\begin{minipage}[t]{0.24\linewidth}
%\centering
%\includegraphics[width=4.7cm]{222.eps}
%
%\end{minipage}%
%}
%\subfigure{
%\begin{minipage}[t]{0.24\linewidth}
%\centering
%\includegraphics[width=4.7cm]{333.eps}
%
%\end{minipage}
%}
%\subfigure{
%\begin{minipage}[t]{0.25\linewidth}
%\centering
%\includegraphics[width=4.7cm]{444.eps}
%
%\end{minipage}
%}
%\centering
%\caption{Comparison of PR curves between our method and the 14 latest methods.}
%\label{fig5}
%\end{figure*}

Boosting Loss aims to enhance the complementarity between branches. As shown in \figref{fig3}, the forward propagation process will activate all branches and obtain $N$ predicted maps $\{P_{i}|i=1,2,3,...,N\}$. The random number $X$ is used to determine which branch performs error back propagation, and the prediction errors of other branches are regarded as weight. In this way, each branch will pay attention to the mispredicted parts of other branches. The calculation process of weight can be expressed as:
\begin{align}
W_{x}=\sum_{i=1}^{N}\mathcal{L}_{bce}(P_{i},g)(i\neq X)+1,
\end{align}
where $g$ represents the ground-truth map. $\mathcal{L}_{bce}$ denotes the pixel-level binary cross-entropy error, which can be expressed as:
\begin{align}
\mathcal{L}_{bce}(p,g)=-((g\otimes log(p))\oplus (\overline{g}\otimes log(\overline{p}))) ,
\end{align}
where $p$ denotes predicted map. $log$ represents a pixel-by-pixel logarithmic operation. $\overline{g}$ and $\overline{p}$ denote the inversion operation pixel by pixel. The boosting loss can be calculated according to the calculated weight. It can be expressed as:
\begin{align}
\mathcal{L}_{b}(p,g,w)= \mathcal{L}_{wbce}(p,g,w)+\mathcal{L}_{wiou}(p,g,w),
\end{align}
where $\mathcal{L}_{wbce}$ and $\mathcal{L}_{wiou}$ represent weighted binary cross entropy loss and weighted intersection of union loss. They can be expressed as:
\begin{align}
\mathcal{L}_{wbce}(p,g,w)= \frac{\mathcal{S}(\mathcal{L}_{bce}(p,g)\otimes w)}{\mathcal{S}(w)},
\end{align}
\begin{align}
\mathcal{L}_{wiou}(p,g,w)=1-\frac{\mathcal{S}(p\otimes g\otimes w)}{\mathcal{S}((p\oplus g)\otimes w)-\mathcal{S}(p\otimes g\otimes w)},
\end{align}

where $\mathcal{S}$ represents the operation of summing all pixels. Other symbols are consistent with the previous description.

In general, MHB decomposes the training process of multi-scale modules so that multiple branches receive training with random samples and relieves the mutual influence of the error back propagation process. BL further enhances the complementarity between each branch. The experiments prove that MHB and BL can effectively improve SOD performance.

\subsection{Loss Function}We utilize the sum of Binary Cross Entropy (BCE) and Intersection of Union (IoU) as the loss function, which is widely used in LDF \cite{LDF}, etc. In addition to the final prediction map, we will also supervise the output features of each Convolution Block (CB) in \figref{fig4}. The total loss function can be expressed as:
\begin{align}
\mathcal{L}_{tot}=\sum_{i=1}^{M}\mathcal{L}_{bi}(F_{i},g)+\mathcal{L}_{b}(P_{4},g,W_{x}),
\end{align}
where $F_{i}$ represents the prediction map of each CB module, and $P_{4}$ represents the final result. $M$ denotes the number of CB modules. $\mathcal{L}_{bi}$ can be expressed as:
\begin{align}
\mathcal{L}_{bi}(p,g)=\mathcal{L}_{wbce}(p,g,W_{1})+\mathcal{L}_{wiou}(p,g,W_{1}),
\end{align}
where $W_{1}$ represents a matrix with all $1$ values, which means that the auxiliary loss is not weighted.

\section{Experiment}

\subsection{Experiment setting} 
The evaluation indicators are mainly as follows: Mean Absolute Error (MAE), max and mean F-measure ($F_{\beta }^{*},F_{\beta }^{m}$) \cite{yang2013saliency}, Maximum enhanced-alignment measure ($E_{\xi}$) \cite{fan2018enhanced}, structure measure ($S_{m}$) \cite{S-measure} and precision-recall (PR). 

The experiment involves the following datasets: DUT-OMRON \cite{yang2013saliency} (5168), ECSSD \cite{yan2013hierarchical} (1000), HKU-IS \cite{li2015visual} (4447), PASCAL-S \cite{li2014secrets} (850), DUTS-TE \cite{wang2017learning} (5019), DUTS-TR \cite{wang2017learning} (10553). We choose DUTS-TR as the training set and other datasets as the test set.
 \begin{table*}[t]
 \caption{Quantitative comparison table with the latest methods on multiple indicators, including the max and mean F-measure ($F_{\beta }^{*}$ and $F_{\beta }^{m}$ the larger the better), MAE (the smaller the better), E-measure ($E_{\xi}$, the larger the better), and S-measure ($S_{m}$, the larger the better) . The best and second best results are marked in {\color{red}{red}}, and {\color{blue}{blue}}, respectively.}
 \renewcommand{\arraystretch}{1.15}
 \setlength\tabcolsep{1pt}
\resizebox{\textwidth}{!}{
\begin{tabular}{c|c c c c c|c ccc c|ccc c c| c ccc c |c ccc c}
\toprule
 & \multicolumn{5}{c|}{\textbf{ECSSD (1000)}} & \multicolumn{5}{c|}{\textbf{HKU-IS (4447)}} & \multicolumn{5}{c|}{\textbf{DUTS-TE (5019)}} & \multicolumn{5}{c|}{\textbf{DUT-OMRON (5168)}} & \multicolumn{5}{c}{\textbf{PASCAL-S (850)}} \\

 \textbf{Method}& $F_{\beta }^{*}\uparrow$ &$F_{\beta }^{m}\uparrow$ & $mae\downarrow$ & $E_{\xi}\uparrow$& $S_{m}\uparrow$& $F_{\beta }^{*}\uparrow$ &$F_{\beta }^{m}\uparrow$ & $mae\downarrow$ & $E_{\xi}\uparrow$& $S_{m}\uparrow$& $F_{\beta }^{*}\uparrow$ &$F_{\beta }^{m}\uparrow$ & $mae\downarrow$ & $E_{\xi}\uparrow$& $S_{m}\uparrow$& $F_{\beta }^{*}\uparrow$ &$F_{\beta }^{m}\uparrow$ & $mae\downarrow$ & $E_{\xi}\uparrow$& $S_{m}\uparrow$& $F_{\beta }^{*}\uparrow$ &$F_{\beta }^{m}\uparrow$ & $mae\downarrow$ & $E_{\xi}\uparrow$& $S_{m}\uparrow$ \\
\midrule

$BASNet_{19}$ &
  .942 &
  .880 &
  .037 &
  .921 &
  .916 &
  
  .928 &
  .895 &
  .032 &
  .946 &
  .909 &
  
  .860 &
  .791 &
  .048 &
  .884 &
  .866 &
  
  .805 &
  .756 &
  .056 &
  .869 &
  .836 &
  
  .857 &
  .775 &
  .076 &
  .847 &
  .832\\

$PoolNet_{19}$ &
  .944 &
  .914 &
  .039 &
  .924 &
  .922 &
  
  .933 &
  .896 &
  .032 &
  .949 &
  .910 &
  
  .880 &
  .811 &
  .040 &
  .889 &
  .878 &
  
  .808 &
  .746 &
  .056 &
  .863 &
  .828 &
  
  .869 &
  .823 &
  .074 &
  .850 &
  .847\\

$CPD_{19}$ &
.939 &
.917 &
.037 & 
.925 &
.918 &

.925 &
.891 &
.034 &
.944 &
.905 &

.865 &
.805 &
.043 &
.887 &
.869 &

.797 &
.747 &
.056 & 
.866 &
.825 &

.864 &
.824 &
.072 & 
.849 &
.842\\

$BANet_{19}$ &
  .945 &
  .880 &
  .035 &
  \color{blue}{.928} &
  .916 &
  
  .931 &
  .895 &
  .032 &
  .950 &
  .909 &
  
  .872 &
  .791 &
  .040 &
  .892 &
  .866 &
  
  .803 &
  .756 &
  .059 &
  .860 &
  .836 &
  
  .870 &
  .775 &
  .070 &
  .855 &
  .832\\
$GateNet_{20}$ &
  .945 &
  .916 &
  .040 &
  .924 &
  .920 &
  
  .933 &
  .899 &
  .033 &
  .949 &
  .915 &
  
  .888 &
  .807 &
  .040 &
  .889 &
  .885 &
  
  .818 &
  .746 &
  .055 &
  .862 &
  .838 &
  
  .875 &
  .825 &
  .068 &
  .852 &
  .852\\
$U2Net_{20}$ &
  .951 &
  .892 &
  .033 &
  .924 &
  .928 &
  
  .935 &
  .896 &
  .031 &
  .948 &
  .916 &
  
  .873 &
  .792 &
  .045 &
  .886 &
  .874 &
  
  .823 &
  .761 &
  .054 &
  .871 &
  .847 &
  
  .862 &
  .772 &
  .076 &
  .841 &
  .836\\

$DFI_{20}$ &
  .949 &
  .920 &
  .035 &
  .924 &
  .927 &
  
  .934 &
  .902 &
  .031 &
  .951 &
  .920 &
  
  .886 &
  .814 &
  .039 &
  .892 &
  .887 &
  
  .818 &
  .752 &
  .055 &
  .865 &
  .839 &
  
  \color{blue}{.885} &
  .837 &
  .065 &
  .857 &
  .861\\

$MINet_{20}$ &
  .947 &
  .924 &
  .033 &
  .927 &
  .925 &
  
  .935 &
  .909 &
  .029 &
  .953 &
  .919 &
  
  .884 &
  .828 &
  .037 &
  .898 &
  .884 &
  
  .810 &
  .755 &
  .055 &
  .865 &
  .833 &
  
  .865 &
  .835 &
  .064 &
  .852 &
  .851 \\

$GCPANet_{20}$ &
  .948 &
  .919 &
  .035 &
  .920 &
  .927 &
  
  .938 &
  .898 &
  .031 &
  .949 &
  .920 &
  
  .888 &
  .817 &
  .038 &
  .891 &
  .891 &
  
  .812 &
  .748 &
  .056 &
  .860 &
  .839 &
  
  .876 &
  .833 &
  .061 &
  .850 &
  .861\\

$ITSDNet_{20}$ &
  .947 &
  .895 &
  .034 &
  .927 &
  .925 &
  
  .934 &
  .899 &
  .031 &
  .952 &
  .917 &
  
  .883 &
  .804 &
  .041 &
  .895 &
  .885 &
  
  .821 &
  .756 &
  .061 &
  .863 &
  .840 &
  
  .876 &
  .792 &
  .064 &
  .853 &
  .856\\

    $LDF_{20}$ &
  .950 &
  .930 &
  .034 &
  .925 &
  .924 &
  
  .939 &
  .914 &
  .027 &
  .954 &
  .919 &
  
  .898 &
  .855 &
  .034 &
  .910 &
  .892 &
  
  .820 &
  .773 &
  .051 &
  .873 &
  .838 &
  
  .874 &
  \color{blue}{.843} &
  \color{blue}{.059} &
  \color{blue}{.865} &
  .856 \\
  
  $PFSNet_{21}$ &
  .952 &
  \color{blue}{.932} &
  \color{blue}{.031} &
  \color{blue}{.928} &
  .930 &
  
  \color{blue}{.943} &
  .919 &
  \color{blue}{.026} &
  \color{blue}{.956} &
  .924 &
  
  .896 &
  .847 &
  .036 &
  .902 &
  .892 &
 
  .823 &
  .774 &
  .055 &
  .875 &
  .842 &
  
  .875 &
  .837 &
  .063 &
  .856 &
  .854 \\
  
      $VST_{21}$ &
  .951 &
  .920 &
  .033 &
  .918 &
  \color{blue}{.932} &
  
  .942 &
  .900 &
  .029 &
  .953 &
  \color{blue}{.928} &
  
  .890 &
  .818 &
  .037 &
  .892 &
  .896 &
  
   .825 &
  .756 &
  .058 &
  .861 &
  .850 &
  
  .875 &
  .829 &
  .061 &
  .837 &
  \color{blue}{.865} \\
  
    $PAKRN_{21}$ &
  \color{blue}{.953} &
  .931 &
  .032 &
  .924 &
  .928 &
  
  \color{blue}{.943} &
  \color{blue}{.920} &
  .027 &
  .955 &
  .923 &
  
  \color{blue}{.907} &
  \color{blue}{.865} &
  \color{blue}{.033} &
  \color{blue}{.916} &
  \color{blue}{.900} &
  
  \color{blue}{.834} &
  \color{blue}{.793} &
  \color{blue}{.050} &
  \color{red}{.885} &
  \color{red}{.853} &
  
  .873 &
  .838 &
  .066 &
  .857 &
  .852 \\
\midrule
  $\textbf{OURS}$ &
  \color{red}{.964} &
  \color{red}{.949} &
  \color{red}{.022} &
  \color{red}{.932} &
  \color{red}{.941} &
  
  \color{red}{.953} &
  \color{red}{.936} &
  \color{red}{.020} &
  \color{red}{.965} &
  \color{red}{.933} &
  
  \color{red}{.920} &
  \color{red}{.890} &
  \color{red}{.025} &
  \color{red}{.925} &
  \color{red}{.910} &

  \color{red}{.838} &
  \color{red}{.804} &
  \color{red}{.040} &
  \color{blue}{.878} &
  \color{blue}{.847} &
  
  \color{red}{.890} &
  \color{red}{.862} &
  \color{red}{.050} &
  \color{red}{.866} &
  \color{red}{.867} \\
\bottomrule
\end{tabular}
}
\label{tab1}
\end{table*}

%  \begin{table*}[t]
%  \caption{The comparison of quantitative results includes the maximum F-measure ($F_{\beta }^{*}$, the larger the better), MAE (the smaller the better) and E-measure ($E_{\xi}$, the larger the better). The best two results are shown in {\color{red}{red}}, and {\color{blue}{blue}}, respectively.}
%  \label{tab:com}
% \resizebox{\textwidth}{!}{
% \begin{tabular}{l|c|c|c|c|c|c|c|c|c|c|c|c|c|c|c}
% \toprule
% \multicolumn{1}{c|}{--}  & \multicolumn{3}{c|}{ECSSD} & \multicolumn{3}{c|}{HKU-IS} & \multicolumn{3}{c|}{DUTS-TE} & \multicolumn{3}{c|}{DUT-OMRON} & \multicolumn{3}{c}{PASCAL-S} \\
% \hline
% \multicolumn{1}{c|}{Method} & $F_{\beta }^{*}\uparrow$ & $mae\downarrow$ & $E_{\xi}\uparrow$&$F_{\beta }^{*}\uparrow$ & $mae\downarrow$ & $E_{\xi}\uparrow$&$F_{\beta }^{*}\uparrow$ & $mae\downarrow$ & $E_{\xi}\uparrow$&$F_{\beta }^{*}\uparrow$ & $mae\downarrow$ & $E_{\xi}\uparrow$&$F_{\beta }^{*}\uparrow$ & $mae\downarrow$ & $E_{\xi}\uparrow$ \tabularnewline
% \midrule
% Ours &0.952 &0.031 &0.928 &0.943 &0.026 &0.956 &0.898 &0.036 &0.902 &0.823 &0.055 &0.875 &0.881 &0.063 &0.857 \tabularnewline
% Ours1 &0.952 &0.031 &0.928 &0.943 &0.026 &0.956 &0.898 &0.036 &0.902 &0.823 &0.055 &0.875 &0.881 &0.063 &0.857 \tabularnewline
% Ours2 &0.952 &0.031 &0.928 &0.943 &0.026 &0.956 &0.898 &0.036 &0.902 &0.823 &0.055 &0.875 &0.881 &0.063 &0.857 \tabularnewline
% Ours3 &0.952 &0.031 &0.928 &0.943 &0.026 &0.956 &0.898 &0.036 &0.902 &0.823 &0.055 &0.875 &0.881 &0.063 &0.857 \tabularnewline
% \bottomrule
% \end{tabular}
% }
We use NVDIA GTX 2080Ti to train our network. The input resolution of the detail branch is $352\times352$, and the input resolution of the first transformer block is $56\times56$. The maximum learning rate of the backbone is 0.004, and the other parts are expanded by ten times. And the learning rate will first increase and then decrease when training our model. The optimization method uses Stochastic Gradient Descent. Batch size is set to 26, and the epoch is set to 32. The data augmentation methods involves multi-scale training, random flipping and cropping. The prediction results do not need any post-processing.
\subsection{Methods of Comparison}The experimental process involves the state-of-the-art methods of the last three years. Four of them in 2019 include BANet \cite{su2019banet}, BASNet \cite{Qin_2019_CVPR}, PoolNet \cite{Liu2019PoolSal} and CPD \cite{CPD}. Seven methods in 2020, including LDF \cite{LDF}, MINet \cite{pang2020multi}, GCPANet \cite{chen2020global}, GateNet \cite{zhao2020suppress}, DFI \cite{liu2020dynamic}, ITSDNet \cite{zhou2020interactive}
and U2Net \cite{qin2020u2}. And there methods published in 2021: PAKRN \cite{PAKRN}, VST \cite{vst} and PFSNet \cite{Ma_Xia_Li_2021}. The evaluation process uses a unified evaluation code to evaluate the published saliency map.
\begin{table}[htb]
\caption{Ablation experiments of the proposed method. Bi is the abbreviation of Bilateral. The last line is the final result of our method.}
 \renewcommand{\arraystretch}{1.1}
 \setlength\tabcolsep{4pt}
\centering
\resizebox{1.0\columnwidth}{!}{
\begin{tabular}{c|cc|cc|cc}
\toprule
 {}& \multicolumn{2}{c|}{\textbf{ECSSD}} & \multicolumn{2}{c|}{\textbf{DUTS-TE}} & \multicolumn{2}{c}{\textbf{DUT-OMRON}}\\ 
 \textbf{Method}& $F_{\beta }^{m}\uparrow$& $mae\downarrow$  & $F_{\beta }^{m}\uparrow$ & $mae\downarrow$ &  $F_{\beta }^{m}\uparrow$ & $mae\downarrow$ \\ \midrule
Detail branch & .867 & .061 & .716 & .067 & .644 & .094 \\
Semantic branch & .848 & .033 & .845 & .033 & .780 & .050 \\
Bilateral & .930 & .027 & .858 & .029 & .788 & .047 \\ \midrule
Bi+AF & .940 & .025 & .869 & .028 & .790 & .045 \\ 
Bi+AF+MBD2 & .941 & .025 & .874 & .027 & .798 & .045 \\ 
Bi+AF+MBD3 & .946 & .024 & .882 & .028 & .799 & .044 \\ 
Bi+AF+MBD4 & .947 & .023 & .888 & .026 & .801 & .041 \\ 
Bi+AF+MBD5 & .946 & .024 & .886 & .027 & .800 & .042 \\
Bi+AF+ASPP & .941 & .025 & .864 & .028 & .793 & .044 \\ \midrule
Bi+AF+MBD4+BL & \textbf{.949} & \textbf{.022} &\textbf{.890} & \textbf{.025} & \textbf{.804} & \textbf{.040} \\
\bottomrule
\end{tabular}
}
\label{tab2}
\end{table}
\subsection{Performance Comparison.}The verification results prove that the proposed method achieves a breakthrough in performance compared with the latest method. First of all, it can be seen from the PR curve in \figref{fig5} that the PR curve of the previous methods has not improved significantly, but our method is significantly better than other methods. In addition, the performance comparison results in \tabref{tab1} can also verify the effectiveness of the method in this paper. Our method has mostly surpassed the previous method in five commonly used indicators. And the performance improvement is very significant. Taking the ECSSD dataset as an example, the MAE indicator reached $0.033$ in 2020, and it has been reduced to $0.031$ in 2021, a reduction of only $6\%$. But our method reached $0.022$, an decrease of \textbf{$29\%$} compared to last year. The experimental data on DUTS-TE shows that the MAE metric of LDF \cite{LDF} in 2020 has reached $0.034$, while the result in 2021 is basically unchanged. But our method reduces the MAE indicator to $0.025$, which is a $24\%$ reduction. From the results of F-measure, the proposed method greatly improves the F-measure value calculated at different thresholds. 
\begin{figure*}[t]
\centering
\includegraphics[width=0.97\textwidth]{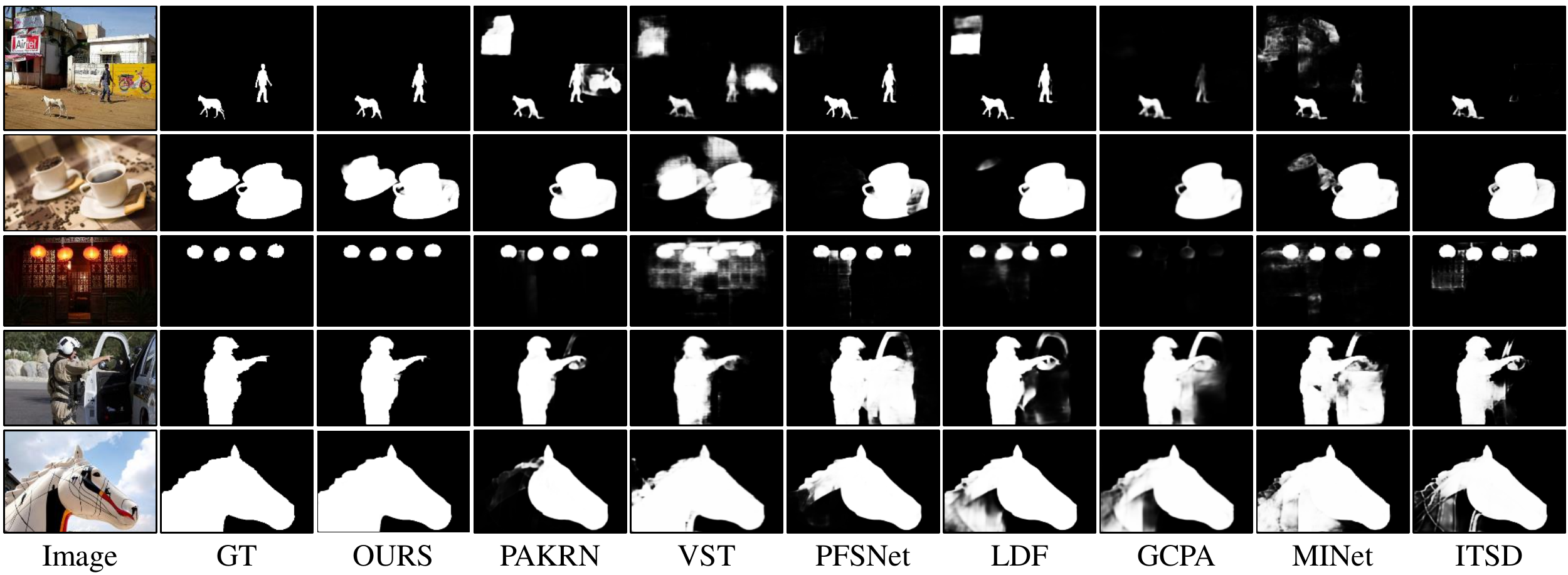} % Reduce the figure size so that it is slightly narrower than the column.
\caption{Visual comparison between the proposed method and the state-of-the-art methods.}
\label{fig7}
\end{figure*}

 \begin{figure}[t]
\centering
\includegraphics[width=0.95\columnwidth]{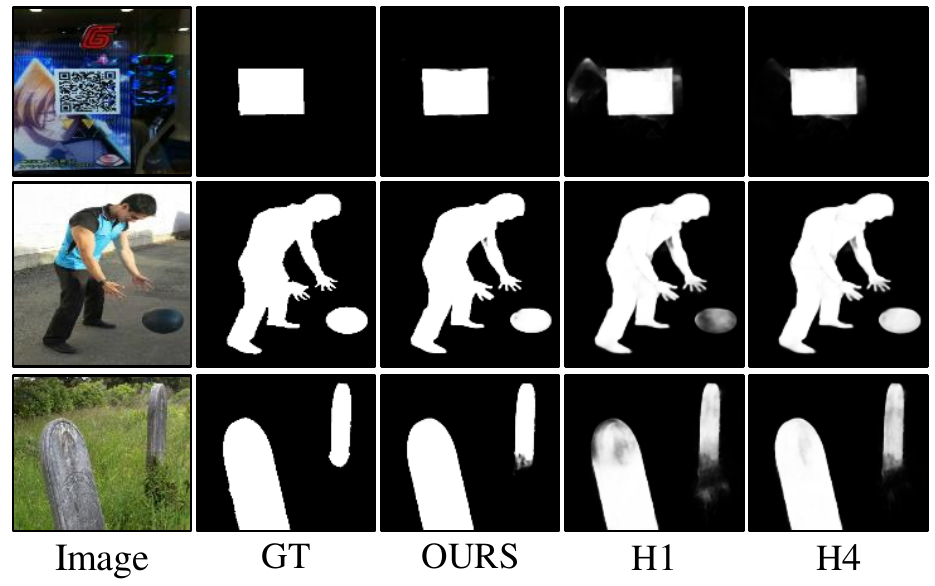} % Reduce the figure size so that it is slightly narrower than the column. Don't use precise values for figure width.This setup will avoid overfull boxes.
\caption{Visualization of prediction results of different prediction heads in MBD. H1, H4 represent the predicted maps of the first and fourth branches, respectively.}
\label{fig6}
\end{figure}

\subsection{Ablation Studies}
In order to verify the innovations of this paper one by one, we conduct ablation experiments on the proposed Bilateral Network, Attention Fusion Module AF, Multi-Branch Decoupling, and Boosting Loss. The experimental details are shown in \tabref{tab2}.
\subsubsection{Bilateral Network.}
The results of the first three rows of \tabref{tab2} verify the effectiveness of the bilateral network. The detailed branch represents a framework composed of a lightweight CNN and FPN \cite{FPN} decoder. Here, the CNN network selects ResNet-18 \cite{ResNet}, and the input resolution is set to $352\times 352$. The semantic branch selects Swin Transformer \cite{liu2021swin} as basic network and utilizes the same decoder. The input resolution of semantic branch is set to $56\times 56$. It can be seen from the experimental results that the performance of a pure CNN or transformer model is not ideal. The lightweight CNN lacks deep semantic information and the transformer model with low-resolution input lacks high-frequency details. When we combine the two models into a bilateral network, the performance can be greatly improved.
\subsubsection{Attention Fusion Module (AF)}
 AF improves their performance through cross-attention compensation. The performance of the fourth row in \tabref{tab2} is significantly improved than that of the third row. On different datasets, the F-measure metric can be improved by more than 1 percentage point. And the Average Absolute Error shows a downward trend. Experiments have proved that the bilateral network can achieve higher performance under the blessing of AF.
\subsubsection{Multi-Branch Decoupling (MBD).}
\tabref{tab2} also verifies the effect of multi-branch decoupling and includes an ablation experiment of the number of branches $N$. The data in the fourth row to the eighth row shows the change in performance as the number of branches increases. When the number of branches increases to 4, the performance is close to saturation. Therefore, we finally chose to use four branches. In addition, the effect of directly adding ASPP to the bilateral network in the table is not obvious. This verifies the role of MBD in the serial training of multi-branch sub-networks.

\subsubsection{Boosting Loss (BL).}In order to verify the effect of BL, we analyze it from both quantitative and qualitative perspectives. The last row in \tabref{tab2} shows the results of BL acting on the 4 branches. Compared to the model without BL, the average performance has improved and basically exceeded the best result of the ablation test. \figref{fig6} shows the visualization results of the first and fourth branches. It is easy to see that different branches can extract different detailed features, and the fusion result can effectively integrate the results of different branches. 
\subsubsection{Visual comparison.}The latest method and the visualization results of the proposed method are shown in \figref{fig7}. It is easy to see that the existing methods mainly have the following problems: 1. Some objects are missing in the prediction result when there are multiple salient objects. 2. Incomplete prediction of a single object. 3. Improper handling of details generates noise. Among the methods compared, VST is based on the transformer model, while other methods are based on CNN. We observe that VST can handle problems 1 and 2 better, but the segmentation details are often not ideal.As shown in the second and fourth rows of the figure, the CNN model is difficult to deal with problem 1 and problem 2 due to the limited receptive field. It can be seen from the third column that our model can effectively combine the advantages of the two types of models. This also verifies the expansion and boosting effect of our method on the receptive field. The results of the first two rows verify that the method in this paper can effectively solve the problem of missing objects. The results in the third and fourth rows show that our method has advantages in detail processing. And the data in the last row shows that even if the object is large, our model can still fully predict and accurately segment the results. In short, the visualization results reflect that the proposed method can better alleviate the problems caused by the receptive field.
\section{Conclusion}
In this paper, we rethink the role of the receptive field in SOD and propose a bilateral network based on vision transformer and lightweight CNN to broaden the receptive field of the network. in order to better integrate the characteristics of the bilateral network, we design an attention feature fusion module based on the characteristics of the two types of features. In the proposed bilateral network, transformer branches input low-resolution images to efficiently generate semantic information. However, this operation causes the global attention mechanism of transformer branch to degenerate into an association mechanism between pixel regions. To enhance the attention between these regions, we propose a Multi-Head Boosting strategy to compensate for the loss of the global receptive field. Experiments show that our method can achieve impressive results under various evaluation indicators on multiple benchmark datasets.
%\subsection{Acknowledgement}

\bibliography{my}
\end{document}